\documentclass[10pt,twocolumn,letterpaper]{article}

\usepackage{iccv}              % To produce the CAMERA-READY version
\usepackage{bm}% To produce the REVIEW version
\usepackage{marvosym}
\usepackage{pifont}
\newcommand{\xmark}{\ding{55}}
\definecolor{iccvblue}{rgb}{0.21,0.49,0.74}
\usepackage[pagebackref,breaklinks,colorlinks,allcolors=iccvblue]{hyperref}

\def\paperID{*****} % *** Enter the Paper ID here
\def\confName{ICCV}
\def\confYear{2025}

\title{VQualA 2025 Challenge on Image Super-Resolution Generated Content Quality Assessment: Methods and Results}

\author{
\makebox[\textwidth][c]{
\begin{tabular}{ccccc}
Yixiao Li\textsuperscript{*} & Xin Li\textsuperscript{*} & Chris Wei Zhou\textsuperscript{*} & Shuo Xing & Hadi Amirpour \\
Xiaoshuai Hao & Guanghui Yue & Baoquan Zhao & Weide Liu & Xiaoyuan Yang  \\
Zhengzhong Tu & Xinyu Li & Chuanbiao Song & Chenqi Zhang & Jun Lan \\
Huijia Zhu & Weiqiang Wang & Xiaoyan Sun & Shishun Tian & Dongyang Yan \\
Weixia Zhang & Junlin Chen & Wei Sun & Zhihua Wang &Zhuohang Shi \\
Zhizun Luo & Hang Ouyang & Tianxin Xiao & Fan Yang & Zhaowang Wu \\
&&Kaixin Deng&&
\end{tabular}
}
}
\makeatletter
\renewcommand\@makefnmark{}
\renewcommand\@makefntext[1]{\noindent#1}
\makeatother
\begin{document}
\maketitle
%\textsuperscript{*}
\footnotetext{\textsuperscript{*}Yixiao Li (18335310648@163.com), Xin Li (xin.li@ustc.edu.cn), and Chris Wei Zhou (zhouw26@cardiff.ac.uk) are the challenge organizers. Shuo Xing, Hadi Amirpour, Xiaoshuai Hao, Guanghui Yue, Baoquan Zhao, Weide Liu, Xiaoyuan Yang, and Zhenzhong Tu are the technical supporters of the challenge. (Corresponding author: \textit{Chris Wei Zhou}).

The other authors are participants of the VQualA 2025 Challenge on Image Super-Resolution Generated Content Quality Assessment.}
\footnotetext{VQualA webpage: \url{https://vquala.github.io/}}
\footnotetext{The ISRGen-QA dataset: \url{https://github.com/Lighting-YXLI/ISRGen-QA}}
\begin{abstract}
This paper presents the ISRGC-Q Challenge, built upon the Image Super-Resolution Generated Content Quality Assessment (ISRGen-QA) dataset, and organized as part of the Visual Quality Assessment (VQualA) Competition at the ICCV 2025 Workshops. Unlike existing Super-Resolution Image Quality Assessment (SR-IQA) datasets, ISRGen-QA places a greater emphasis on SR images generated by the latest generative approaches, including Generative Adversarial Networks (GANs) and diffusion models. The primary goal of this challenge is to analyze the unique artifacts introduced by modern super-resolution techniques and to evaluate their perceptual quality effectively. A total of 108 participants registered for the challenge, with 4 teams submitting valid solutions and fact sheets for the final testing phase. These submissions demonstrated state-of-the-art (SOTA) performance on the ISRGen-QA dataset. The project is publicly available at: \url{https://github.com/Lighting-YXLI/ISRGen-QA}.
\end{abstract}    
\section{Introduction}
\label{sec:intro}
Super-resolution (SR) image quality assessment metrics aim to evaluate the perceptual quality of SR images from a human-centric perspective. This necessity arises from the inherently ill-posed nature of the SR task, where a single low-resolution image may correspond to multiple plausible high-resolution reconstructions. As a result, SR faces a fundamental challenge in balancing fidelity (i.e., similarity to the ground truth) and naturalness (i.e., perceptual realism)~\cite{sfsn,srif}. Moreover, SR images exhibit distinct distortion characteristics that differ significantly from those found in traditionally degraded images. Conventional distortions (e.g., blur, noise, and compression artifacts)~\cite{zhou2025perceptual} typically stem from information loss and lead to perceptual degradation. In contrast, SR methods often introduce enhancement-induced artifacts, including over-sharpened edges, hallucinated or false textures, and unnatural reconstruction patterns. Consequently, accurately assessing the perceptual quality of SR images is crucial~\cite{li2024deep,zhou2020blind}—not only for evaluating but also for guiding the design and optimization of next-generation super-resolution algorithms.

To assess the SR-specific distortions, several SR-IQA datasets have been built, including the QADS~\cite{qads}, Waterloo~\cite{waterloo}, SISR-IQA~\cite{sisriqa}, CVIU~\cite{cviu}, RealSRQ~\cite{realsrq}, and SISAR~\cite{SISAR}. The QADS~\cite{qads} database contains 20 original HR references selected from the MDID database~\cite{SUN2017153}, and 980 SR images created by 21 SR algorithms, including 4 interpolation-based, 11 dictionary-based, and 6 DNN-based SR models, with upsampling factors equaling 2, 3, and 4. Each SR image is associated with the mean opinion score (MOS) collected from 100 individuals. In the CVIU~\cite{cviu} database, 1,620 SR images are produced by 9 SR approaches from 30 HR references. The HR reference images are selected from BSD200 according to the PSNR values. Six pairs of scaling factors (i.e., 2, 3, 4, 5, 6, 8) and kernel widths (i.e., 0.8, 1.0, 1.2, 1.6, 1.8, 2.0) are adopted, where a larger subsampling factor corresponds to a larger blur kernel width. Each image is rated by 50 subjects, and the mean of the median 40 scores is calculated for each image as the MOS. The Waterloo~\cite{waterloo} database involves 13 original HR references at 512×512 resolution, 39 low-resolution (LR) references, and 312 interpolated SR images generated by 8 interpolation algorithms, with upsampling factors of 2, 4, and 8. Subjective scores were collected from 30 participants aged 20–30 (17 males and 13 females). The SISR-IQA~\cite{sisriqa} database contains 15 ground-truth HR images selected from Set5, Set14, and BSD100, and 360 SR images reconstructed using 8 SR algorithms (e.g., DRCN~\cite{DRCN} and VDSR~\cite{VDSR}) with upsampling factors of 2, 3, and 4. LR images were generated via nearest neighbor interpolation. Subjective quality scores for all 360 SR images were collected from 16 participants, who were unaware of the HR references and SR methods. The RealSRQ~\cite{realsrq} database contains 60 real-world HR references and 180 corresponding LR images at three scaling factors (2, 3, 4). A total of 1,620 SR images were generated using 10 SISR algorithms, including 5 non-deep methods and 5 deep models (SRCNN~\cite{SRCNN}, CSCN~\cite{CSCN}, VDSR~\cite{VDSR}, SRGAN~\cite{SRGAN}, and USRnet~\cite{USRnet}). Subjective scores were collected from 60 participants (32 males and 28 females). The SISAR~\cite{SISAR} database contains 12,600 SR images generated from 100 natural LR images. These images were processed using 10 SR algorithms or combinations, including 2 interpolation-based, 4 learning-based (SRCNN~\cite{SRCNN}, VDSR~\cite{VDSR}, RCAN~\cite{RCAN}, SAN~\cite{SAN}), and 4 hybrid methods (e.g., SRCNN+BICUBIC). SR images were generated at 6 different scaling factors (i.e., 1.5, 2, 2.7, 3, 4, 3.6). Subjective scores were collected from 23 participants aged 20–30 with normal vision. Despite the progress in constructing SR-IQA databases, the super-resolution algorithms employed in these datasets have failed to keep pace with the rapid advancements in the field. For example, the most recent methods included are USRNet~\cite{USRnet} (2020) and SAN~\cite{SAN} (2019), with SRGAN being the only generative adversarial network (GAN)-based approach represented. This lag significantly limits the effectiveness and generalizability of the resulting SR-IQA metrics in evaluating modern SR techniques.

With the rapid advances of the generative methods in SR tasks, recent SR models have increasingly incorporated generative priors, particularly through GAN- and diffusion-based models~\cite{zhang2019ranksrgan,li2024sed,yang2025diffusion,li2025difiisr,moser2025dynamic}. While these methods have shown promising results, striking an effective balance between perceptual realism and reconstruction fidelity remains challenging. GAN-based approaches often yield high-fidelity metrics but may fail to capture vivid textures due to their unstable adversarial training and tendency toward over-optimization~\cite{wu2024seesr}. While diffusion models can produce detailed textures by leveraging powerful generative priors, their reliance on stochastic noise sampling and mismatches between prior and LR distributions can undermine pixel-level accuracy~\cite{yu2024scaling}. Consequently, the construction of subjective quality assessment datasets specifically involving SR images produced by the latest generative models is of paramount importance for facilitating the further refinement and advancement of SR techniques. 

In conjunction with the ICCV 2025 Workshop, we present the ISRGC-Q Challenge on image super-resolution generated content quality assessment. The goal of this challenge is to automatically evaluate the perceptual quality of super-resolved (SR) images, ensuring that the predicted scores align as closely as possible with human visual perception. This challenge is one of the VQualA 2025 Workshop associated challenges on: FIQA: Face Image Quality Assessment Challenge~\cite{ma2025fiqa}, ISRGC-Q: Image Super-Resolution Generated Content Quality Assessment Challenge~\cite{isrgcq2025iccvw}, EVQA-SnapUGC: Engagement Prediction for Short Videos Challenge~\cite{li2025evqa}, Visual Quality Comparison for Large Multimodal Models~\cite{zhu2025vqa}, DIQA: Document Image Enhancement Quality Assessment Challenge~\cite{diqa2025iccvw}, GenAI-Bench AIGC Video Quality Assessment~\cite{genai-bench2025iccvw}. In the following sections, we describe the challenge in detail, present and analyze the results, and provide an overview of the participating methods.

\section{VQualA 2025 Challenge on ISRGC-Q}
\label{sec:formatting}
The VQualA 2025 Challenge on Image Super-Resolution Generated Content Quality Assessment (ISRGC-Q) is the first challenge to be organized to advance the development of assessing super-resolved images, especially those generated by the latest GAN- and diffusion-based approaches. The details of the whole challenge are as follows:
\subsection{ISRGen-QA database}
ISRGen-QA is a super-resolution (SR) image quality assessment database that contains sufficient SR images generated by the latest generative models, including GAN- and diffusion-based methods. It consists of 720 super-resolved images at approximately 2K resolution ($2040 \times 1152\sim2040\times1440$), covering four typical upscaling factors ($\times$2, $\times$3, $\times$4, and $\times$8). A total of 15 advanced SR algorithms are used to generate the images, including 4 GAN-based (i.e., ESRGAN~\cite{esrgan}, Real-ESRGAN~\cite{realesrgan}, BSRGAN~\cite{bsrgan}, and SeD~\cite{li2024sed}), 5 diffusion-based (i.e., SR3~\cite{SR3}, IDM~\cite{IDM}, SRDiff~\cite{srdiff}, CDFormer~\cite{cdformer}, and SAM-DiffSR~\cite{samdiffsr}), 4 transformer-based (i.e., SRNO~\cite{srno}, ATD-SR~\cite{ATD}, SwinIR~\cite{swinir}, and CAMixerSR~\cite{camixersr}), 1 flow-based method (i.e., BFSR~\cite{bfsr}), and 1 CNN-based method (i.e., EDSR~\cite{edsr}). The SR images are derived from 19 high-resolution (HR) reference images and 76 low-resolution (LR) reference images created via four down-sampling scales ($\times$2, $\times$3, $\times$4, and $\times$8). The HR references are selected from the DIV2K~\cite{div2k}, and the utilized down-sampling method is the Bicubic. To ensure the reliability of perceptual quality annotations, subjective scores were collected from 23 human participants (11 female, 12 male, from 5 different countries and various ages), with anomaly filtering yielding valid scores from 21 participants. The dataset is divided into training (576 images, $80\%$), validation (72 images, $10\%$), and test (72 images, $10\%$) sets, facilitating reproducible model development and benchmarking.
\subsection{Evaluation Protocol}
This challenge utilized two metrics to measure the correlation of the quality predictions and the mean opinion scores (MOS), including Spearman rank-order correlation coefficient (SRCC) and Pearson linear correlation coefficient (PLCC). SRCC and PLCC are employed to assess the monotonicity and accuracy of predictions, respectively. An ideal quality metric would have SRCC and PLCC values close to one. The final score used for ranking is computed by reweighting the above metrics as :
\begin{equation}
    \rm{Score} = 0.6\times \rm{SRCC} + 0.4\times\rm{PLCC}.
\label{eq1}
\end{equation}
\subsection{Challenge Phases}
There are two phases in this challenge, i.e., the development and testing phases. The details are as follows:
\subsubsection{Development Phase:}
In the development phase, we release 576 SR images and their corresponding high-resolution and low-resolution reference images in our ISRGen-QA dataset to support each team in developing their algorithms. Moreover, we release 72 SR images without their MOS scores for validation. Each participant can upload their quality predictions of the validation set to the challenge platform (Codalab: \url{https://codalab.lisn.upsaclay.fr/competitions/22924}). Then they can obtain the corresponding final score,
SRCC, and PLCC. In the development phases, we received 193 submissions from 12 teams in total.
\subsubsection{Testing Phase:}
In the testing phases, we release 72 SR images and their corresponding high-resolution and low-resolution reference images in our ISRGen-QA dataset for testing. The final ranking is achieved with
the score in Eq. \ref{eq1}. In the test stage, 5 teams submitted their final results to the challenge platform. At the end of this competition, we received the fact sheets and source codes
from 4 teams, which are utilized for the final ranking.
\section{Challenge Results}
% We have summarized the challenge results in Table 1.

The main results from the 4 participating teams (Team MICV, Team ydy, Team QA-Veteran, and Team 2077 Agent) are summarized in Table \ref{tab:results}, as well as the detailed information on their methods. Figures \ref{fig:results-hist} and \ref{fig:results} present illustrations of the performance achieved by the submitted methods.

\subsection{Results Analysis}

As presented in Table \ref{tab:results}, all participating teams demonstrated exceptional performance, achieving overall scores exceeding 0.9, which indicates strong alignment with human perceptual quality judgments. A consistent pattern emerges across all teams: PLCC scores consistently exceed SRCC values (ranging from 0.9476 to 0.9714 for PLCC versus 0.9096 to 0.9588 for SRCC). This discrepancy reveals important insights about the methods' characteristics:

\begin{itemize}
    \item \textbf{Linear relationship capture:} The higher PLCC scores indicate that all methods excel at capturing linear correlations between predicted and ground truth quality scores.
    
    \item  \textbf{Rank-order consistency:} The relatively lower SRCC scores suggest some challenges in maintaining perfect monotonic rank consistency across the entire quality spectrum.

    \item \textbf{Practical implications:} While methods may occasionally misorder samples with similar quality levels, they maintain strong overall quality prediction accuracy.
\end{itemize}

\begin{table*}[htbp]
\centering
\footnotesize
\caption{Quantitative results from the VQualA 2025 Image Super-Resolution Generated Content Quality Assessment Challenge, including detailed information on the methods used by the 4 participating teams. The best performances are highlighted in bold. Note that \textbf{GFlops} are calculated relative to \textbf{Input Size}.}
\setlength{\tabcolsep}{3.pt}
\renewcommand{\arraystretch}{1.5}
% \small
\begin{tabular}{c|l|l|ccc|ccc|c|c}
\toprule
\textbf{Rank} & \textbf{Team} & \textbf{Leader} & \textbf{Overall} & \textbf{SRCC} $\uparrow$ & \textbf{PLCC} $\uparrow$ & \textbf{Params. (M}) & \textbf{Input Size} & \textbf{GFlops (G)} & \textbf{Ensemble} & \textbf{Extra Data} \\
\midrule
% Baseline             &  &  &  &  &  \\
% \hline
1 & MICV & Chuanbiao Song & \textbf{0.9638} &  \textbf{0.9588} & \textbf{0.9714} & 6 & (448, 448, 3) &1000 & \xmark & \xmark \\
2 & ydy & Shishun Tian & 0.9429 & 0.9333 & 0.9572  & 161 & (128, 128, 3) & 6 & \xmark & \xmark\\
3 & QA-Veteran & Weixia Zhang &  0.9409 & 0.9277 & 0.9608 & 375.32 & (1280, 1280, 3) & 428.83 & \xmark & \xmark\\
4 & 2077 Agent & Zhuohang Shi & 0.9248  & 0.9096 & 0.9476 & 91.56 & (2040, 1152, 3) & 322.73 & \xmark & \xmark\\
\bottomrule
\end{tabular}
\label{tab:results}
\end{table*}

\begin{figure}[t]
    \centering
    \includegraphics[width=1.\linewidth]{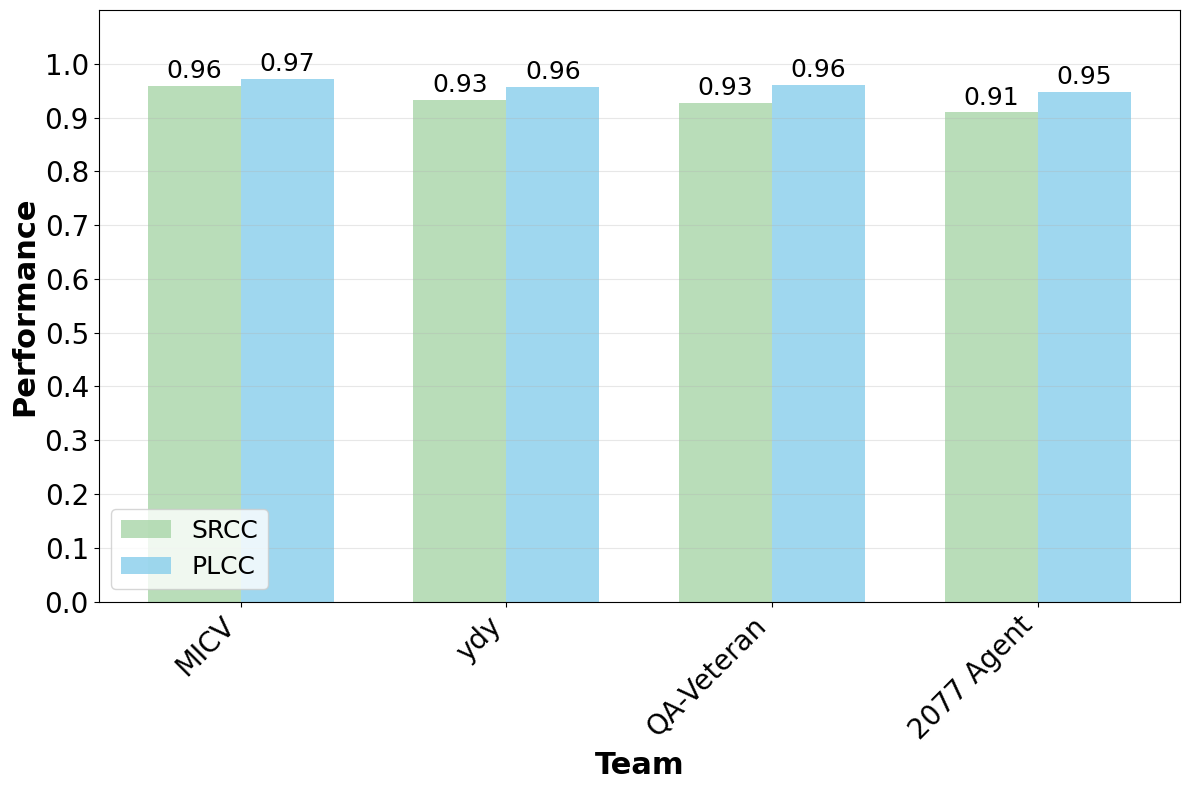}
    \caption{The performance of the methods submitted by different teams on the testing set.}
    \label{fig:results-hist}   
    % \vspace{-5mm}
\end{figure}

Figure \ref{fig:results-hist} shows the performance histogram comparing PLCC and SRCC scores across the 4 participating teams. The consistently high performance scores, tightly clustered between 0.91 and 0.97, highlight the effectiveness of the submitted super-resolution quality assessment methods.

\begin{figure*}[h]
    \centering
    \includegraphics[width=0.24\linewidth]{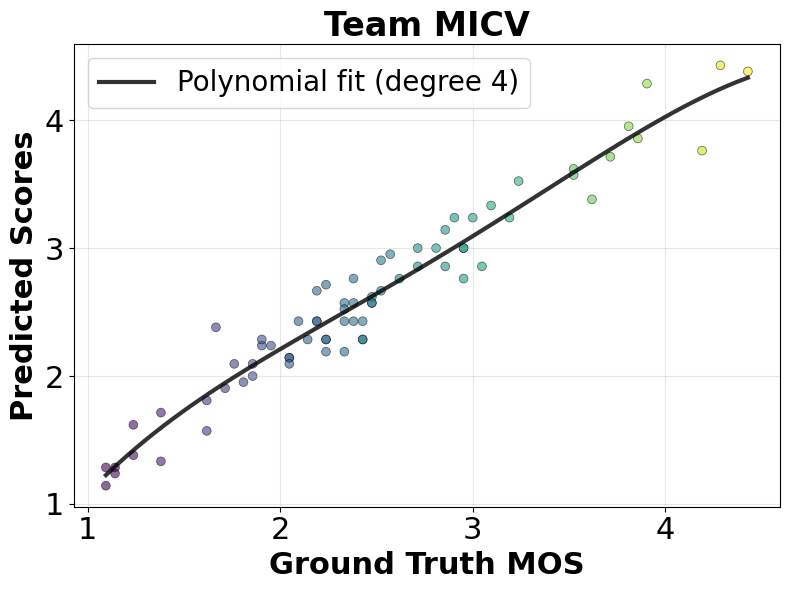}
    \includegraphics[width=0.24\linewidth]{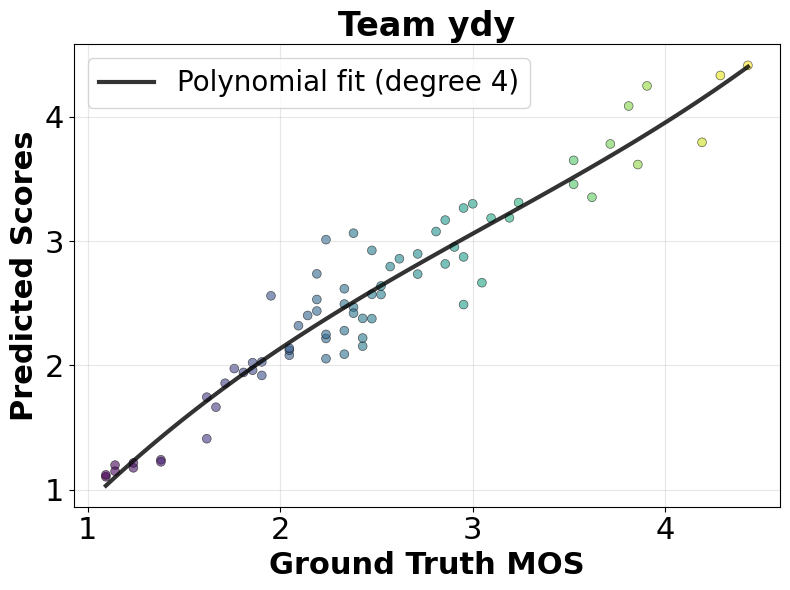}
    \includegraphics[width=0.24\linewidth]{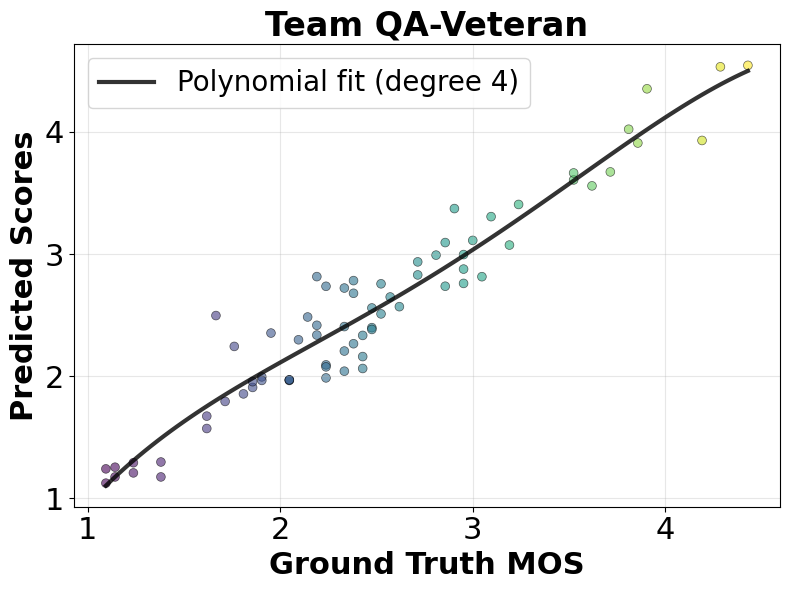}
    \includegraphics[width=0.24\linewidth]{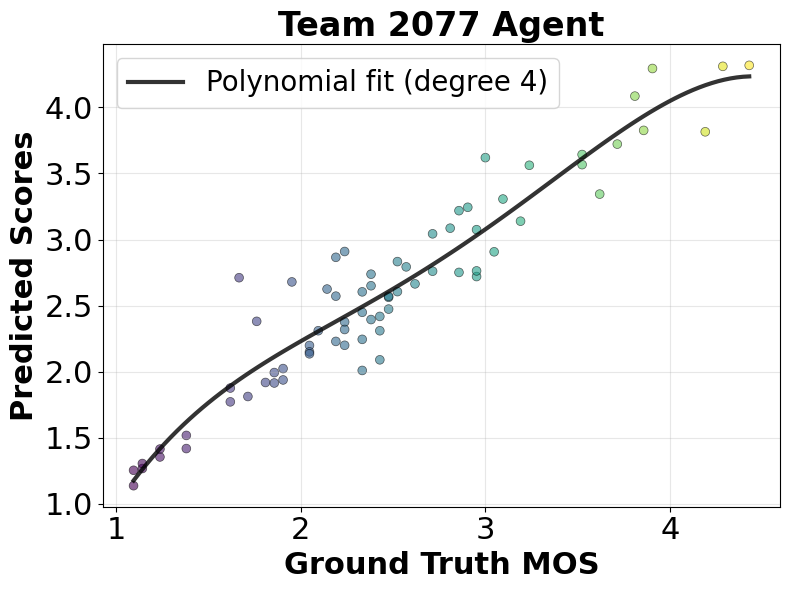}
    \caption{Scatter plots of predicted scores versus MOS for all participating teams on the testing set. The curves are obtained by a fourth-order polynomial nonlinear fitting.}
    \label{fig:results}   
    % \vspace{-5mm}
\end{figure*}

Furthermore, Figure \ref{fig:results} shows the scatter plots of predicted scores versus the MOS for all 4 team methods on the testing set. The curves are obtained through fourth-order polynomial nonlinear fitting. We can observe that the predicted scores obtained by the top-performing team methods demonstrate higher correlations with the MOS values, as evidenced by the tighter clustering of data points around the fitted curves.

% Overall, all participating teams demonstrated exceptional performance, achieving overall scores exceeding 0.9, which indicates strong alignment with human perceptual quality judgments. 

\section{Teams and Methods}
\label{sec:rationale}
\subsection{MICV Team}
\begin{figure}[t]
\centering
\includegraphics[width=\linewidth]{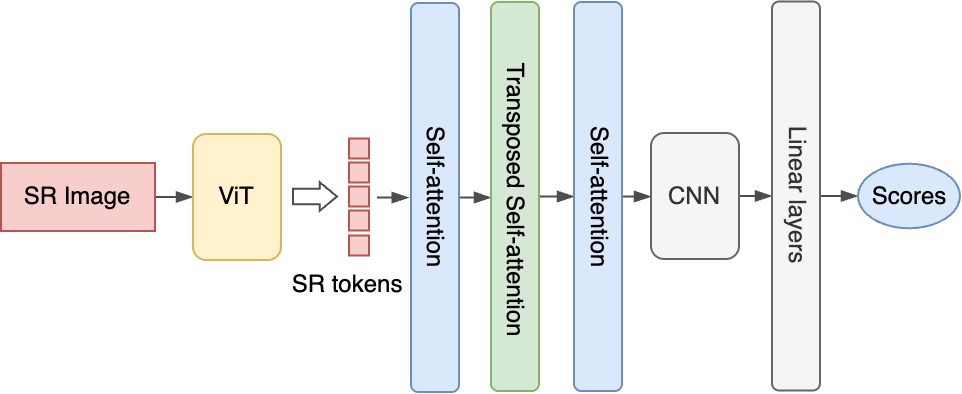}
\caption{Model architecture of the proposed Hybrid Vision Transformer and Convolutional Neural Network for Super-Resolution Image Quality Assessment.}
\label{VqualA}
\end{figure}

The MICV team proposes the Hybrid Vision Transformer (ViT) and Convolutional Neural Network (CNN) for SR Image Quality Assessment~\cite{xinyu2025hybrid}. The proposed method focuses solely on the SR images as input, leveraging the ViT to capture global dependencies and the CNN to extract spatial features. The architecture of the method is illustrated in Figure \ref{VqualA}. To adaptively model the visual features of SR images, they introduce a multi-stage attention mechanism that enhances hierarchical feature fusion through self-attention and transposed self-attention.

Specifically, the ViT takes the SR image as input and generates high-level image tokens. These tokens are first processed by a self-attention module to capture global contextual relationships. Subsequently, the output is transposed and fed into a second self-attention layer, enabling the model to learn dependencies along alternative spatial directions. The output of the transposed self-attention is then reverted to its original dimensionality and passed through a third self-attention layer, further refining cross-scale interactions. Finally, the enriched feature map is encoded through multiple convolution layers, linear layers, and a sigmoid activation function to predict the quality score. 

By eliminating the need for LR \& HR priors, the model achieves a more streamlined architecture while maintaining strong performance in quality estimation. This design not only reduces computational complexity and GPU memory, but also avoids potential biases introduced by reference image dependencies.

%Furthermore, they have explored the generalizability of Multimodal Large Language Models (MLLMs) to this task, hypothesizing that their strong contextual modeling capabilities could enhance prediction accuracy. However, experimental results revealed suboptimal performance of MLLMs in low-level regression tasks. Analysis indicates that MLLMs exhibit limited capacity for local texture perception and global structural awareness essential for super-resolution image quality assessment. Moreover, their high parameter count significantly increases training costs. In contrast, lightweight models based on ViT-CNN hybrid architectures demonstrated superior efficiency and stability in SR image quality prediction tasks. This experience underscores that for low-level visual tasks, specifically designed compact models may outperform general-purpose MLLMs in terms of both effectiveness and computational efficiency.

\paragraph{Training Setup} This task aims to learn models to predict the MOS of 21 participants for SR images. The training data consists of 576 SR images from the ISRGen-QA dataset, with their corresponding low- and high-resolution counterparts, covering upscaling levels of ×2, ×3, ×4, and ×8. During the data pre-processing phase, the SR images are directly used as input, with no reference to LR or HR counterparts. Data augmentation is implemented through random horizontal flipping and random cropping operations to enhance model robustness. 
The resulting cropped images maintain a resolution of 448×448 pixels, with any entirely black images being re-cropped to ensure data validity. 
The AdamW optimizer is employed with an initial learning rate of 1e-5, and learning rate scheduling is conducted using cosine annealing. 
The loss function was defined as a weighted sum of the PLCC loss and SRCC loss with equal weighting ratios of 1:1. 
The training is conducted on 8 NVIDIA A100 GPUs with a batch size of 4 for 200 epochs.

\paragraph{Testing Details} During the validation phase, 72 SR images are utilized to evaluate the effectiveness of the trained model. Following the same pre-processing procedure as in the training phase, only the SR images are processed, with no involvement of LR or HR counterparts. Subsequently, center cropping is performed to extract 448×448 resolution images, which are then directly input into the network for quality score prediction.
Subsequently, center cropping is performed to extract input resolution images, which are then processed by the network for quality score prediction.

\subsection{ydy Team}
\begin{figure}[t]
\centering
\includegraphics[width=\linewidth]{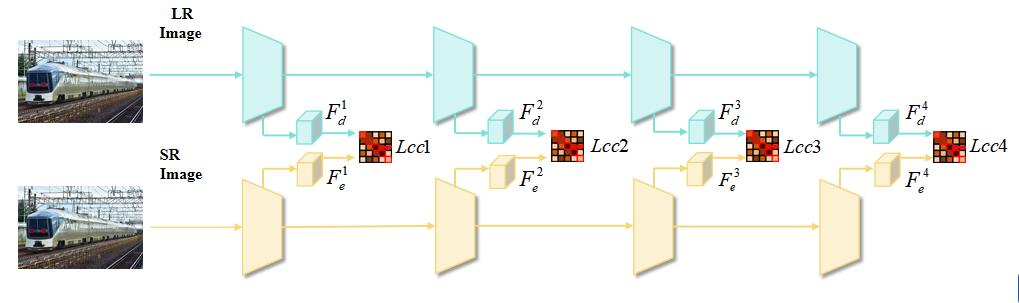}
\caption{Model architecture of the proposed Cross-Covariance Loss Calculation in team ydy.}
\label{ydy2}
\end{figure}
\begin{figure}[t]
\centering
\includegraphics[width=\linewidth]{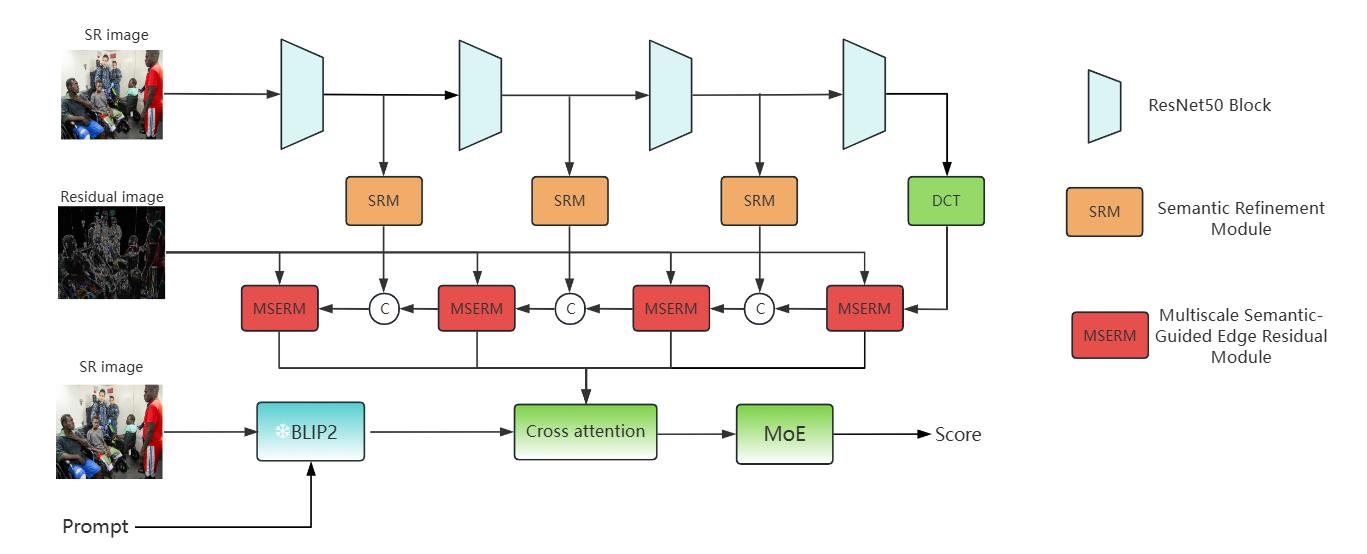}
\caption{Model framework of the proposed BLIP-2 Assisted Residual-Guided Quality Assessment for Super-Resolution Images.}
\label{ydy1}
\end{figure}
\begin{figure}[t]
\centering
\includegraphics[width=\linewidth]{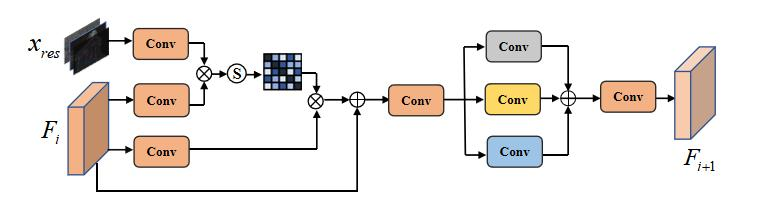}
\caption{Model framework of the proposed SRM: Semantic Refinement Module in team ydy.}
\label{ydy3}
\end{figure}
\begin{figure}[t]
\centering
\includegraphics[width=\linewidth]{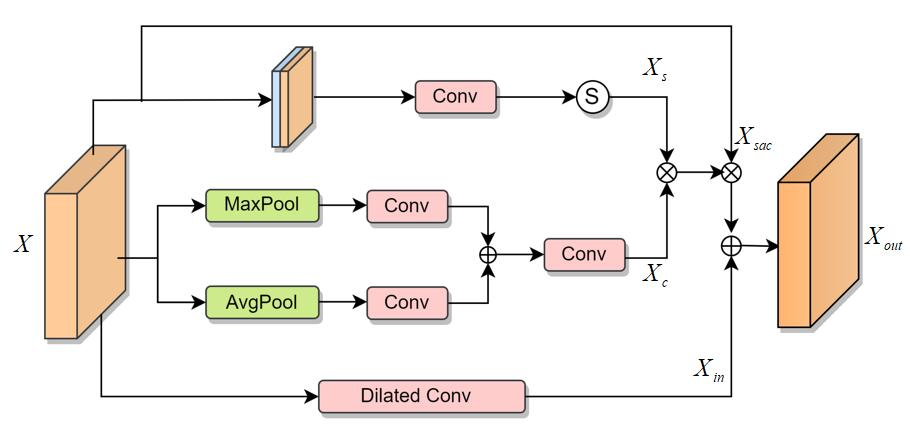}
\caption{Model framework of the proposed MSERM: Multiscale Semantic-Guided Edge Residual Module in team ydy.}
\label{ydy4}
\end{figure}
The ydy team proposes the BLIP-2 Assisted Residual-Guided Quality Assessment for Super-Resolution Images~\cite{tian2025blip2}, which is a hybrid quality assessment network tailored for SR images, integrating semantic information from BLIP-2 features and residual guidance from precomputed residual maps. The architecture builds upon a dual-branch backbone with ResNet50 encoders for both super-resolved and low-resolution images. The model is designed to mitigate SR-specific artifacts, such as edge distortions and texture inconsistencies, through multi-level residual guidance and semantic alignment in SR-IQA tasks. First, hierarchical features from the SR image and its corresponding low-quality reference are extracted using two parallel ResNet50 backbones, as shown in Figure \ref{ydy2}. Their feature-level similarity is supervised via a cross whitening loss, ensuring consistency at multiple depths. The residual image—capturing high-frequency differences between SR and LR—is processed through a multi-stage pooling operation to generate hierarchical residual cues. These cues are fused with the backbone features via a series of channel-wise Self-Attention blocks, allowing residual signals to guide perceptual feature enhancement, as shown in Figure \ref{ydy1}.  The Semantic Refinement Module (SRM, as shown in Figure \ref{ydy3}) enhances semantic representation between the encoder and decoder by integrating spatial, channel, and semantic cues. It captures global context via dilated convolution and refines features through spatial and channel attention, whose outputs are fused and combined with semantic features to guide decoding. And the Multiscale Semantic-guided Edge Residual Module (MSERM, as shown in Figure \ref{ydy4}) leverages semantic-aware edge residuals to refine features using an attention mechanism, followed by multiscale enhancement via parallel dilated convolutions. Together, SRM and MSERM ensure semantically rich, edge-sensitive, and context-aware feature representations. Finally, a Mixture-of-Experts (MoE) gating mechanism is applied, where a set of projection experts is weighted dynamically based on the gated attention output to produce the final quality prediction.

%Additionally, they incorporate a DCT-based spectral attention mechanism to emphasize frequency-domain discrepancies, particularly on the deepest convolutional layer. This spectral feature is combined with spatial self-attention, enabling the network to capture both structural and frequency-related distortions. A fusion path aggregates multi-level features using MAF (Multi-scale Attention Fusion) modules and L2 pooling for local contrast normalization. These features are projected and normalized through fully connected layers. To integrate semantic information from the BLIP-2 multi-modal model, a cross-attention module fuses the semantic features extracted by BLIP-2 with residual-based structural cues to enhance perceptual representation. 
\paragraph{Training Setup} The proposed model was implemented using PyTorch 2.0.0 and Python 3.9, and trained on a single NVIDIA RTX 3090 GPU. The training was conducted for 100 epochs with a batch size of 16, requiring approximately 8 hours in total. They used the official ISRGC-Q training dataset, which includes 576 SR images along with their corresponding LR and HR reference images and MOS as supervision signals. For each image, 30 training and 30 test patches were randomly sampled. Semantic features extracted from BLIP-2 were precomputed and stored as `.pt` files for training. The optimization was performed using Adam with a fixed learning rate of 1e-4 and a weight decay of 5e-4. During training, they applied standard data augmentation techniques including random horizontal flipping, random cropping to 128×128 patches, and normalization using ImageNet statistics. The model was trained with a multi-objective loss function, combining an L1 loss between predicted scores and MOS, a Cross-Covariance Loss (CCL) between multilevel SR and LR features, and a cosine similarity loss between the visual-semantic outputs and the precomputed BLIP-2 query vectors.

\paragraph{Testing Details} During testing, the proposed model followed the same patch-wise strategy used during training. Each super-resolved image was divided into 30 patches of size 128×128, and a quality score was predicted for each patch. The final image-level score was obtained by averaging the patch-level predictions. As a test-time preprocessing step, they applied center cropping to extract representative image regions. This helps standardize input distributions and reduce prediction variance. BLIP-2 multimodal features were pre-extracted and loaded during inference. All testing was conducted on an NVIDIA RTX 3090 GPU.

\subsection{QA-Veteran Team}
The QA-Veteran team proposes Blind Super-resolution Quality Assessment based on a Resolution-adaptive Vision-language Model~\cite{zhang2025blind}. Image super-resolution aims to recover a high-resolution (HR) image from an LR input. This is a typical ill-posed problem, as multiple plausible HR outputs may correspond to the same LR input. Therefore, to reliably evaluate the quality of super-resolved images, they argue that a no-reference (NR) or blind IQA approach should be adopted. Compared with natural images (e.g., photos taken by a camera), super-resolved images exhibit two distinct characteristics:
\begin{itemize}

    \item \textbf{Algorithm-dependence}: The quality scores of super-resolved images are highly influenced by the specific super-resolution algorithm used. Therefore, an IQA model must be capable of effectively capturing the degradation artifacts introduced by different algorithms.
    
    \item \textbf{High resolution} (e.g., 2K, 4K, etc.): This poses a challenge when applying conventional image preprocessing methods, such as aggressively downsampling the image before feeding it into the model. Such downsampling may obscure the fine-grained features that are crucial for distinguishing the quality differences between high-resolution super-resolved images, ultimately reducing the effectiveness of the quality assessment.
\end{itemize}

To address the above challenges, the proposed method is built upon \textbf{SigLIP2-NaFlex}~\cite{tschannen2025siglip2multilingualvisionlanguage}. On one hand, the model benefits from large-scale image-text pretraining, which enables it to learn rich image representations and better capture the distortion characteristics introduced by different super-resolution algorithms. On the other hand, the \textbf{NaFlex} mechanism in SigLIP2 preserves the original \textit{aspect ratio} and \textit{resolution} of super-resolved images as much as possible, allowing the model to retain fine-grained quality cues that are crucial for distinguishing between high-resolution super-resolved images.
\begin{figure}
    \centering
\includegraphics[width=\linewidth]{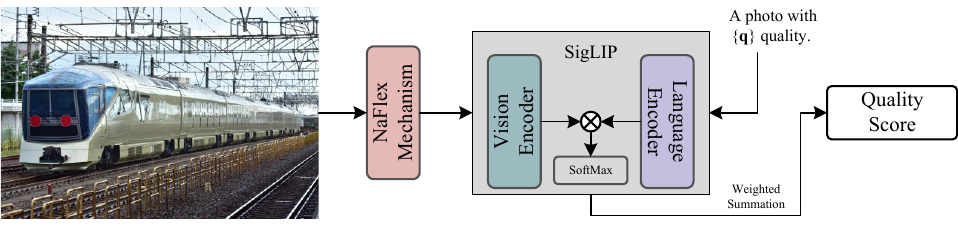}
    \caption{The diagram of the proposed Blind Super-resolution Quality Assessment based on a Resolution-adaptive Vision-language Model.}
    \label{fig:diagram}
\end{figure}

As shown in Figure~\ref {fig:diagram}, given an input super-resolved image $\bm{x}$, they leverage the vision encoder within the SigLIP-2 model to obtain a visual embedding vector. Inspired by~\cite{zhang2023liqe}, they design a textual template: ``\textit{a photo with $\{c\}$ quality}'', where $c \in\mathcal{C}=\{1,2,3,4,5\} =$ \{``bad", ``poor", ``fair”, ``good”, ``perfect"\}. They then use the language encoder within the SigLIP-2 model to encode the textual embedding of all entries in the textual template. They compute the cosine similarity of the visual embedding and all the textual entries and apply a softmax function to obtain the probability distribution of $\bm{x}$ over five quality levels $\hat{p}(c|\bm{x})$. They relate $c$ to a scalar quality score $\hat{q}$ by $\hat{q}(\bm{x}) = \sum_{c=1}^{C}\hat{p}(c|\bm{x})\times c$, where $C = 5$ is the number of quality levels.
% \begin{itemize}

\paragraph{Training Setup} They build their model on the SigLIP2-base-patch16-NaFlex. During training, they randomly choose the maximum number of patches from 4624, 5184, and 5776. They train the model using a single NVIDIA A5880-ada GPU, using the AdamW optimizer~\cite{LoshchilovH19} with a decoupled weight decay regularization of $10^{-3}$. The initial learning rate is set to $5\times10^{-6}$, which is scheduled by a cosine annealing rule~\cite{LoshchilovH17}. They optimize the model for $6$ epochs with a mini-batch size of $12$. They use a combination of fidelity loss~\cite{zhang2021uncertainty}, PLCC loss, and L1 loss as the loss function.

\paragraph{Testing Details} During inference, they fix the maximum number of patches to 4624. The NaFlex mechanism will adaptively preprocess the input image.
% \end{itemize}
\subsection{2077 Agent Team}
The 2077 Agent team proposes the Ultra-High-Resolution Image Quality Assessment\cite{shi2025ultrahighresolution}. SR images refer to images converted from LR to HR through super-resolution reconstruction techniques, with pixel density ranging from tens to hundreds of times that of traditional low-resolution images, thus belonging to the category of ultra-high-resolution images. Taking 8K super-resolution images as an example, their resolution of 33 million pixels significantly enhances the capability of detail representation. However, the quality assessment of super-resolution images faces significant challenges:

\begin{itemize}
    \item \textbf{Massive Data Volume}: The enormous data size of SR images results in prohibitively high computational and storage costs in traditional pixel-level evaluation algorithms, making effective training and deployment difficult.
    \item \textbf{Architectural Limitations}: Existing frameworks based on CNN and Transformers are primarily designed for low-resolution images, struggling to adapt to the scale characteristics of SR images.
\end{itemize}

Traditional high-resolution image quality assessment methods typically employ downsampling or patch-based cropping strategies, but these approaches have notable limitations:
\begin{itemize}
    \item \textbf{Downsampling} inevitably leads to loss of fine details, compromising the accuracy of the assessment.
    \item \textbf{Patch-based Evaluation} divides an image into blocks and calculates an average score as the overall quality metric, implicitly assuming uniform contributions from all blocks. However, this assumption often fails in practice. For instance, when an SR image contains entirely black anomalous blocks, human subjective evaluation would consider such blocks significantly detrimental to overall quality, whereas average scoring underestimates their negative impact, creating a substantial discrepancy between algorithmic results and human perception. Consequently, novel algorithms are urgently needed to address these technical bottlenecks in SR image quality assessment.
\end{itemize}

To overcome these issues, this study breaks away from the conventional MOS evaluation paradigm for entire images. Instead, it dynamically allocates weights based on the varying contributions of different regions to subjective quality perception: texture-rich detail regions are assigned higher weights, while smooth background regions receive lower weights. The final weighted regional evaluations are globally fused via a Score Transformer, generating MOS scores that better align with human visual perception.

Thus, this study proposes \textbf{UltraR-IQA}, an ultra-high-resolution image quality assessment algorithm, as shown in Figure \ref{fig:architecture-grid_1}. By constructing a multi-stage processing system, it enables quantitative quality comparisons between SR images and HR images across multiple spatial scales and abstraction levels. The algorithmic architecture comprises five core modules:
\begin{figure}
    \centering
\includegraphics[width=\linewidth]{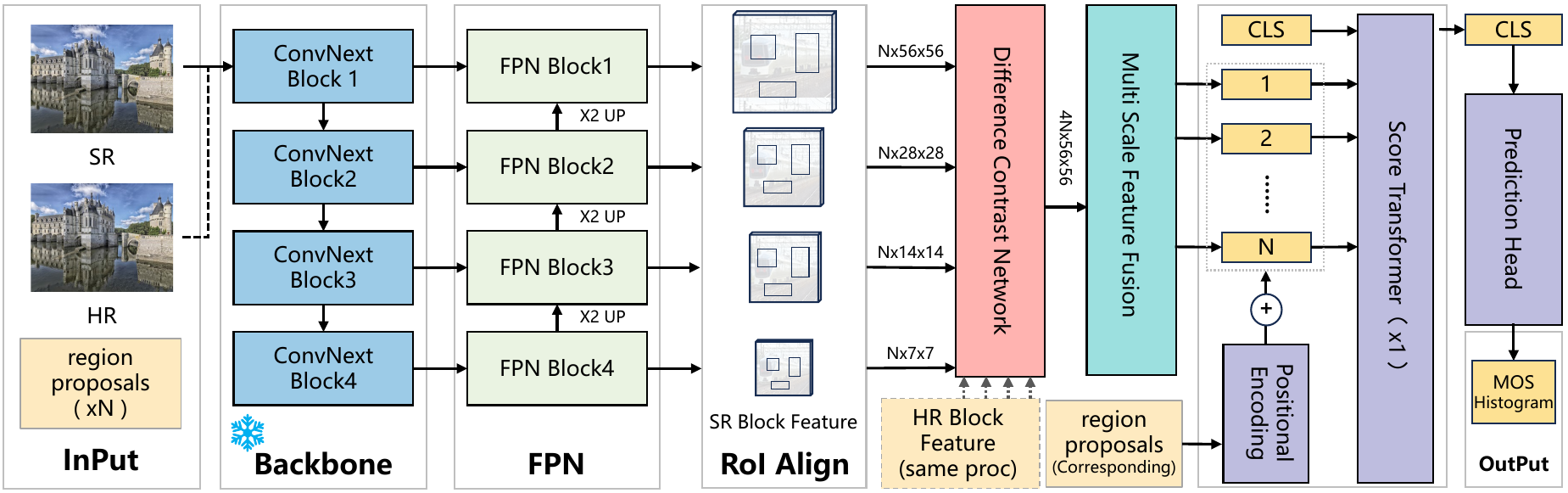}
    \caption{UltraR-IQA Network Architecture: Inputting SR/HR and pre-generated candidate boxes, freezing ConvNext Base backbone parameters, randomly selecting N candidate boxes for ROI Align cropping with the corresponding HR/SR cropping and the same processing; feeding same-layer features into difference Contrast network, fusing cross-layer features via multi-scale network to obtain latent scores, embedding candidate positions with position encoding, and outputting a 5-dimensional frequency distribution.}
    \label{fig:architecture-grid_1}
\end{figure}
\begin{figure}
    \centering
\includegraphics[width=\linewidth]{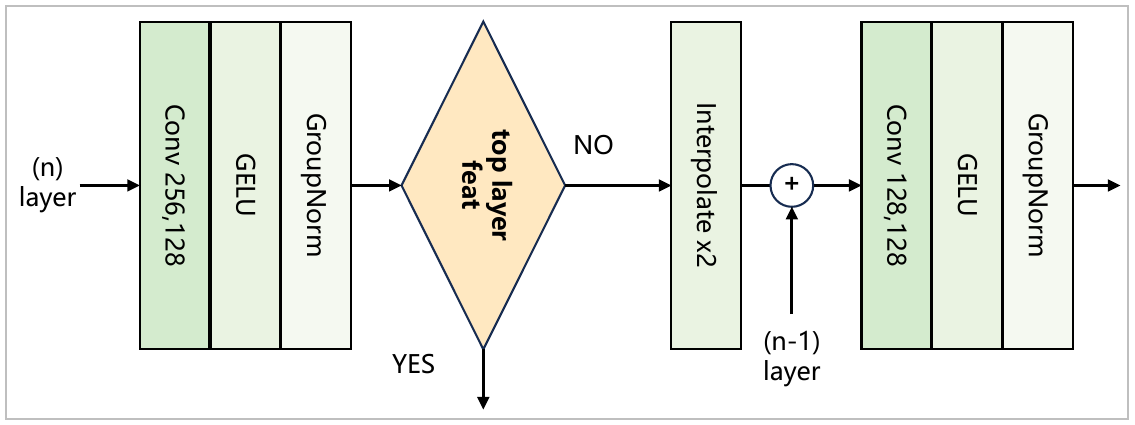}
    \caption{FPN Module Architecture: Taking multi-level ConvNext features as input and outputting 256-channel features; non-highest level inputs undergo upsampling, addition with upper-level features, and subsequent convolution.}
    \label{fig:architecture-grid_2}
\end{figure}
\begin{figure}
    \centering
\includegraphics[width=\linewidth]{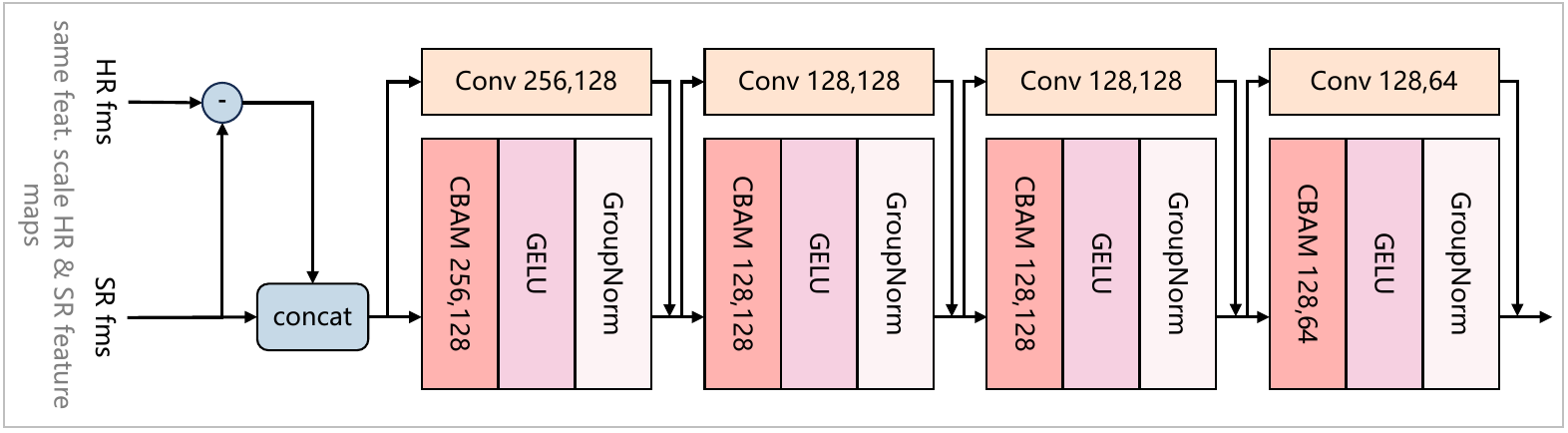}
    \caption{Difference Contrast Network Architecture: Inputting same-layer candidate box features, where ``-" denotes feature subtraction and ``concat" represents channel-wise concatenation.}
    \label{fig:architecture-grid_3}
\end{figure}
\begin{figure}
    \centering
\includegraphics[width=\linewidth]{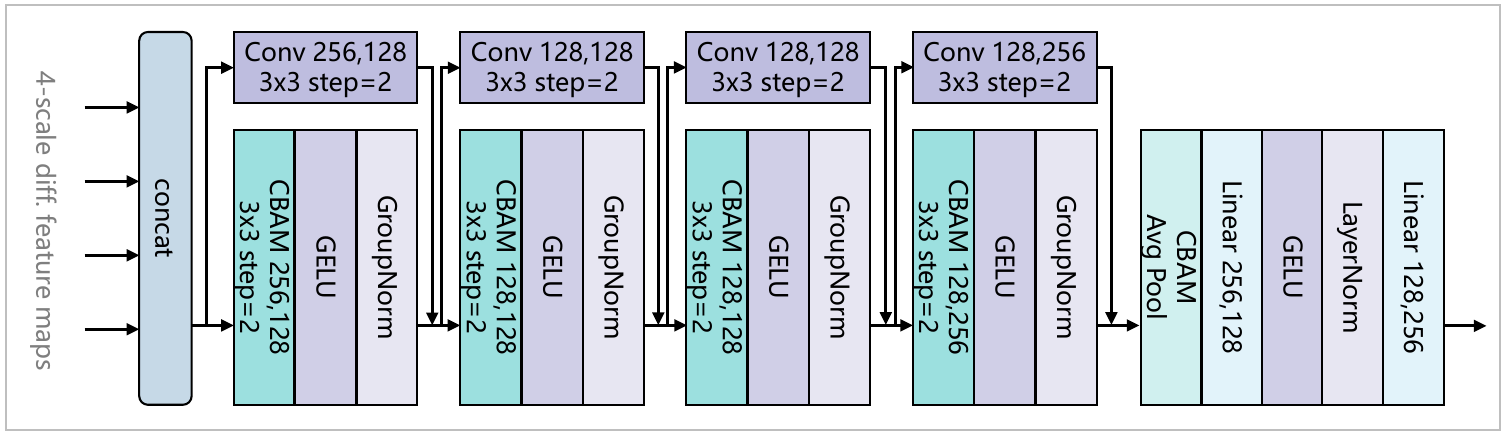}
    \caption{Multi-Scale Feature Fusion Architecture: Employing convolutional downsampling with padding 1 and stride 3; taking multi-scale difference features as input and outputting latent scores for candidate regions.}
    \label{fig:architecture-grid_4}
\end{figure}
\begin{itemize}
    \item \textbf{Candidate Region Generation}: Candidate Region Generation extracts key candidate regions from HR images using the Selective Search algorithm, laying the foundation for subsequent fine-grained analysis due to its ability to generate high-quality candidate regions with varying scales and aspect ratios, crucial for capturing diverse visual features in SR and HR images.
    \item \textbf{Multi-Scale Feature Extraction}: Combines the ConvNeXt-Base backbone network with a Feature Pyramid Network (FPN, as shown in Figure \ref{fig:architecture-grid_2}) to efficiently extract multi-scale features from HR and SR images.
    \item \textbf{Difference Contrast Network}: For each candidate region, a difference contrast network (as shown in Figure \ref{fig:architecture-grid_3}) that integrates residual connections and Convolutional Block Attention Modules (CBAM) precisely computes feature discrepancies.
    \item \textbf{Multi-Scale Feature Fusion}: As shown in Figure \ref{fig:architecture-grid_4}, this module generates regional latent scores through a multi-scale fusion strategy, emphasizing the complementary nature of features at different levels.
    \item \textbf{Score Prediction}: Leverages a Score Transformer model incorporating Fourier and geometric positional encoding to aggregate regional latent scores and output a 5-level MOS frequency histogram distribution, mapping feature differences to subjective quality scores.
\end{itemize}

For scenarios involving gigapixel ultra-high-resolution images with limited local computational resources, this method supports decomposing images into multiple sub-regions for serial processing. By dynamically allocating region-specific weights via an attention mechanism, it achieves efficient and accurate image quality assessment, effectively mitigating the underestimation of outlier impacts caused by average scoring. When computational resources are sufficient, the entire image can also be processed directly, with both approaches yielding approximately equivalent evaluation results. 
% \begin{itemize}
\paragraph{Training Setup} This study pioneers a minimum-variance constrained approach for constructing MOS frequency histograms, identifying scoring combinations that maximize alignment with target MOS values while minimizing variance. According to competition protocols, MOS scores are independently assigned by 21 evaluators. Assuming integer ratings within the 1-5 range, the algorithm employs five-layer nested loops to exhaustively traverse all possible scoring combinations across evaluators, filtering sets where the mean score equals the target value. Subsequently, variance is computed for each qualifying combination, with lower variance indicating more concentrated score distributions. The minimal-variance combination is selected as the optimal solution, forming the foundational data for MOS frequency histogram construction. Computational verification confirms that the constructed MOS frequency histogram exhibits an expectation error of $2.56\times10^{-15}$ relative to ground truth, with each MOS score corresponding to a unique optimal combination. Finally, the model employs a KL-divergence loss function, deriving the ultimate MOS prediction by computing the mathematical expectation of this 5-dimensional MOS frequency histogram.

The framework utilizes a pre-trained ConvNeXt-Base backbone network (supporting Tiny variants). Optimization employs the AdamW optimizer with initial learning rate $1 \times 10^{-4}$ and weight decay coefficient 0.1. Training employs an actual batch size of 1 with gradient accumulation over 4 steps (effective batch size=4) for 100 epochs. Learning rate scheduling follows the StepLR strategy, specifically halving the rate every 25 epochs to enable dynamic optimization.

During image preprocessing, HR and SR images undergo normalization before SR images are overlaid onto HR counterparts using the top-left corner as the alignment origin for dimensional unification. Candidate regions meeting specified criteria are pre-generated via the Selective Search algorithm (persisted before training). Training data augmentation incorporates random horizontal flipping (50\% probability) and 90° interval rotation. All training procedures were executed on a single NVIDIA RTX 2080 Ti GPU.
%\noindent \textbf{Candidate Region Extraction \& Feature Cropping} 
%Employ the Selective Search algorithm to stochastically generate $N$ candidate regions ($N=32$ in this implementation). For each region, precisely crop the corresponding features from both HR and SR images via FPN. Standardize multi-scale features through interpolation to fixed dimensions: $7\times7$, $14\times14$, $28\times28$, and $56\times56$.
%\noindent\textbf{Differential Feature Computation \& Multi-Scale Fusion} 
%Calculate feature discrepancies between HR and SR counterparts at identical scales using shared Difference Contrast Networks. Subsequently, upsample features across scales to $56\times56$ resolution. Integrate cross-scale information through concatenation into Multi-Scale Feature Fusion networks, ultimately generating latent quality scores for candidate regions.
%\noindent\textbf{Dynamic Weight Score Prediction} 
%The latent scores of each candidate region are inputted, and the self-attention mechanism of the Transformer is utilized to adaptively assign weights to different regions. The final MOS of the entire image is obtained through weighted fusion. Additionally, a positional encoding module is introduced, which deeply integrates Fourier features with geometric features to encode the positions of candidate bounding boxes. This operation significantly enhances the model's accuracy in perceiving target locations.
\paragraph{Testing Details} During the inference phase, the operational workflow mirrors the training procedure. The system executes multiple iterations of candidate region sampling and inference operations (default: 20 iterations) based on precomputed candidate regions, subsequently averaging the predictions across iterations to enhance result stability. Furthermore, input images consistently retain their original resolution without modification.
% \end{itemize}
\section{Acknowledgments}
We thank the VQualA 2025 sponsors: Snap Research, INTSIG, and TAOBAO $\&$ TMALL Group.

{
    \small
    \bibliographystyle{ieeenat_fullname}
    \bibliography{main}
}


\end{document}